\documentclass[10pt,twocolumn,letterpaper]{article}

\usepackage{3dv}
\usepackage{times}
\usepackage{epsfig}
\usepackage{graphicx}
\usepackage{amsmath}
\usepackage{amssymb}
\usepackage[table,xcdraw]{xcolor}
\makeatletter
\@namedef{ver@everyshi.sty}{}
\makeatother
\usepackage{tikz}

\usepackage{booktabs}
\usepackage{subfiles}
\usepackage{mathtools}
\usepackage{multirow}

\DeclareMathOperator*{\argmin}{arg\,min}
\usepackage{cuted}
\usepackage{capt-of}
\usepackage{cite}

\usepackage{microtype}
\usepackage{float}

\usepackage{tikz}
\usepackage{comment}
\usepackage{color}

\threedvfinalcopy 

\def\threedvPaperID{****} 

\ifthreedvfinal\fi
\begin{document}

\title{GNPM: Geometric-Aware Neural Parametric Models}

\author{Mirgahney Mohamed\\
University College London\\
\and
Lourdes Agapito\\
University College London\\
}

\maketitle
\thispagestyle{empty}

\begin{strip}
\centering
\includegraphics[width =  \linewidth]{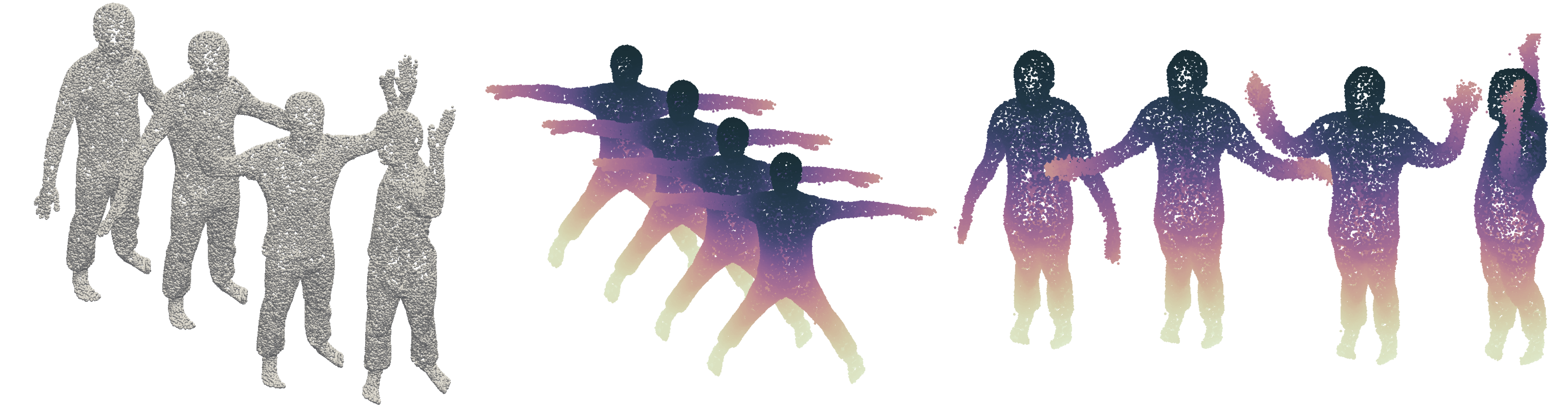}
\captionof{figure}{\textbf{Geometric-Aware Neural Parametric Model:}
            Given a set of input 3D point-clouds of different identities in a variety of poses, GNPM disentangles shape and deformations by mapping each pose to its canonical t-pose and learning dense temporally consistent correspondences, without the need for any ground-truth annotations. }
        \label{fig:teaser}
\end{strip}

\begin{abstract}
   We propose Geometric Neural Parametric Models (GNPM), a learned  parametric  model that takes into account the local structure of  data to learn disentangled shape and pose latent spaces of 4D dynamics, using a geometric-aware architecture on point clouds. Temporally consistent 3D deformations are estimated without the need for dense correspondences at training time, by exploiting cycle consistency. 
   Besides its ability to learn dense correspondences, GNPMs also enable latent-space manipulations such as interpolation and shape/pose transfer. 
   We evaluate GNPMs on various datasets of clothed humans, and show that it achieves comparable performance to state of the art methods that require dense correspondences during training.
\end{abstract}

\section{Introduction}
\label{sec:intro}

Reconstructing temporally consistent shape deformations from visual inputs remains a challenging open problem in computer vision with many applications in AR/VR and content creation. While model-agnostic approaches that exploit local shape smoothness priors~\cite{Newcombe15} have been extremely successful, the remarkable recent advances in machine learning driven capture of domain-specific 3D parametric models such as for human bodies~\cite{SMPL15,Scape05,Joo15}), hands~\cite{MANO17}, faces~\cite{PaysanKARV09, FLAME17, Wang19} or animals~\cite{ZuffiKJB16} have made them an attractive alternative. However, parametric models suffer from two important drawbacks. First, since they do not exploit local geometry priors they often fail to capture local geometry details and have a tendency to over-regularize. Secondly, their construction typically requires manual intervention to aid alignment, or fully dense annotations across all samples to provide dense correspondences at training time.

Neural Parametric Models (NPMs)~\cite{Palafox21} were recently proposed to tackle some of the challenges faced by traditional parametric models by leveraging deep neural networks and implicit functions to learn a disentangled representation of shape and pose. NPMs are learnt from data alone without the need for any class specific knowledge and, once trained, can be fit to new observations via test-time optimization. While NPMs offer an appealing alternative to traditional 3D parametric models, since they only require the same identity to be seen in different poses (including a canonical pose) dropping the need for registration across different identities, they still require known dense correspondences during training between different poses of the same identity and their canonical pose. Moreover, NPMs do not learn any local geometry regularity priors given that flow predictions are conditioned on global shape and pose codes and local geometry priors are not exploited.

To tackle both limitations we propose Geometric Neural Parametric Models (GNPMs). We represent surfaces as point clouds and exploit the ability of dynamic graph neural network architectures~\cite{YueWang18} to explicitly take into account the local structure around deformed points to learn local features and enforce local geometric regularity. In addition, we relax the need for any correspondences at all during training by disentangling 4D dynamics into shape and pose latent spaces via a cycle consistency loss.

Geometric priors are enforced by adapting the EdgeConv~\cite{YueWang18} graph convolution operators to design our model as an auto-decoder~\cite{DeepSDF}. Our intuition is that points tend to deform coherently with their neighbours~\cite{ARAP07}. Furthermore, local features learnt via graph convolutions can also help the model to learn semantic relations between non-neighbouring points, which is potentially useful in deformations such as dancing where distant body parts move synchronously.
Inspired by ~\cite{Paschalidou2021CVPR}, we show that the learned features can be additionally used to segment shapes into semantically meaningful parts which remain consistent across identities, in a completely unsupervised way (see Figure~\ref{fig:deformed_clust}).
While it is known that geometric-based methods can be inefficient ~\cite{Wu19}, we provide an efficient implementation that allows our model to run at a comparable cost to MLP-based models.

To learn without known correspondences we exploit the observation that transformations between posed and canonical spaces should be cycle consistent. This loss allows us to infer a dense deformation field, by jointly learning the weights of two networks that perform a bi-directional mapping between posed and canonical spaces and the respective shape and pose latent spaces (see Fig.~\ref{fig:self_cycle}). To fit the model to new unseen identities and/or deformations, we use test-time optimisation to minimise the cycle consistency loss to recover shape and pose latent vectors. 

In summary, our contributions are:
\begin{itemize}
    \item GNPM is a geometric-aware neural parametric model that learns to disentangle pose and shape exploiting the local structure of the data via edge convolutions. 
    \item GNPM learns shape and pose embeddings and long-term dense correspondences without the need for ground truth annotations during training.
    \item GNPMs learn rich geometric features useful for downstream tasks such as unsupervised part segmentation.
\end{itemize}

\section{Related Work}

\textbf{Parametric Models:}
Parametric 3D models have become a prevalent tool to model deformable 3D shapes.
They learn to disentangle deformations into several factors of interest and have been applied to various domains such as human bodies ~\cite{Scape05, Joo15, SMPL15, GHUMcvpr20}, hands and faces ~\cite{MANO17} ~\cite{PaysanKARV09, FLAME17, Wang19} and animals ~\cite{ZuffiKJB16}.
Despite their success, they struggle to capture fine-grained details like wrinkles or to model clothes. Also, their construction can be tedious as it often requires domain knowledge or manual tuning. Neural based methods~\cite{Palafox21} offer a compelling alternative to learn directly from data without the need for manual tuning or domain-specific knowledge.

\textbf{Supervised Neural Deformation Models:}
Building on 3D OccNet~\cite{Mescheder18}, OFlow~\cite{OccFlowNiemeyer19ICCV} learns a continuous spatio-temporal representation of 4D dynamics that assigns motion vectors to every location in space-time. However, it degrades when capturing long sequences. 
%
Inspired by Dynamic Fusion~\cite{Newcombe15}, Bozic \emph{et al.}~\cite{bozic2021neuraldg} learn a globally consistent deformation graph while learning dense surface details via local MLPs. However, it is limited to be sequence-specific and cannot be used for shape or pose transfer. 
Palafox \emph{et al.}~\cite{Palafox21} recently introduced Neural Parametric Models (NPMs) leveraging the representation power of implicit functions to disentangle latent spaces of shape and pose. However, NPMs disregard the local  geometric structure of 3D shapes and require dense correspondences for training.
While our geometric model also learns to disentangle between shape and pose,  unlike~\cite{Palafox21} we take into account the geometric structure and can learn long term correspondences in a self-supervised manner without the need for dense ground truth annotations.

\textbf{Self-Supervised Neural Deformation Models:}
LoopReg~\cite{Bhatnagar2020} was the first end-to-end learning framework to solve scan registration with a self-supervision loop. Backward and forward maps are learnt to predict correspondences between every input scan point and the model surface. SCANimate~\cite{Saito2021} also uses a self-supervision cycle to learn an implicit dense field of skinning weights to map surface points to a canonical pose. Although both~\cite{Bhatnagar2020, Saito2021} can model shape deformation in a semi-supervised or unsupervised way, they do not learn latent shape/deformation spaces and rely on SMPL as the underlying body model.
In contrast, our model, GNPM, learns disentangled shape and deformation latent spaces. More recently, Neuromorph~\cite{Eisenberger21} jointly solves shape interpolation and dense correspondences in an unsupervised way  using edge convolutions~\cite{YueWang18} and a single forward pass. Although Neuromorph can estimate dense correspondences between shapes of different categories in different poses the latent interpolation is modelled via time, limiting its ability to learn a parameterised representation.

\section{Method}
\begin{figure}[t]
    \centering
    \includegraphics[width = 1\linewidth]{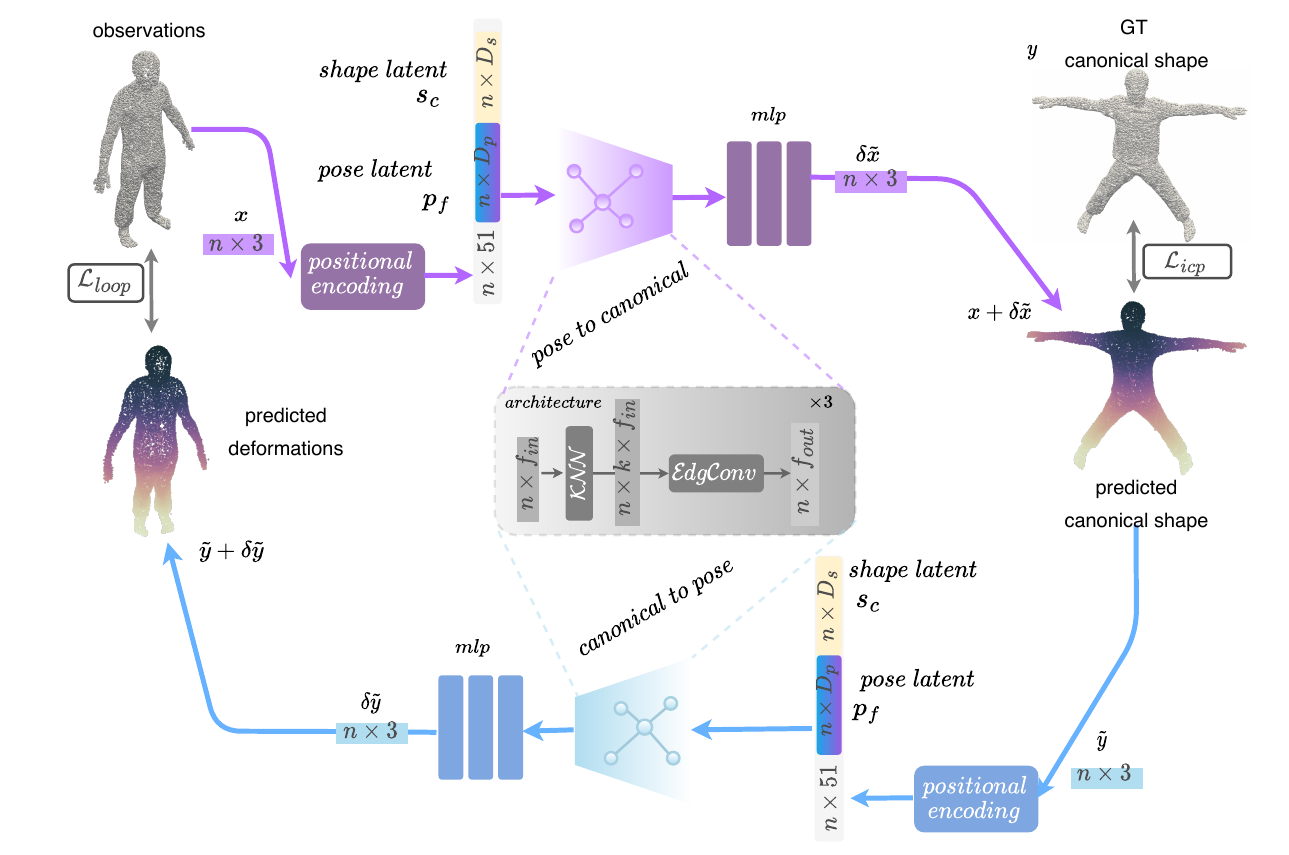}
    \caption{\textbf{Self-supervision cycle.}
    Our model is trained to map an input scan to their canonical t-pose and then back to the deformed shape, imposing cycle consistency. In this way, we learn dense correspondences without any ground truth annotations. In addition, we disentangle shape and pose by learning two embeddings which are jointly optimised with the weights of the graph neural networks.}
    \label{fig:self_cycle}
\end{figure}

We introduce Geometric Neural Parametric Models (GNPM), a geometric-aware model that disentangles 4D dynamics into latent spaces of shape identity and pose and can be learnt without the need for correspondences across shapes.

We choose point clouds as our shape representation given their lightweight nature and that they naturally match the raw output of commodity depth cameras. Moreover, when combined with the EdgeConv architecture~\cite{YueWang18}, local geometric structure can be exploited to capture both local and global shape properties, unlike the more recently popular implicit representations~\cite{Palafox21}.
We adopt EdgeConv layers (see~\ref{back:edge_conv}) and modify the architecture to an auto-decoder~\cite{DeepSDF}. 

\begin{figure*}[t]
    \centering
    \includegraphics[width=1\linewidth]{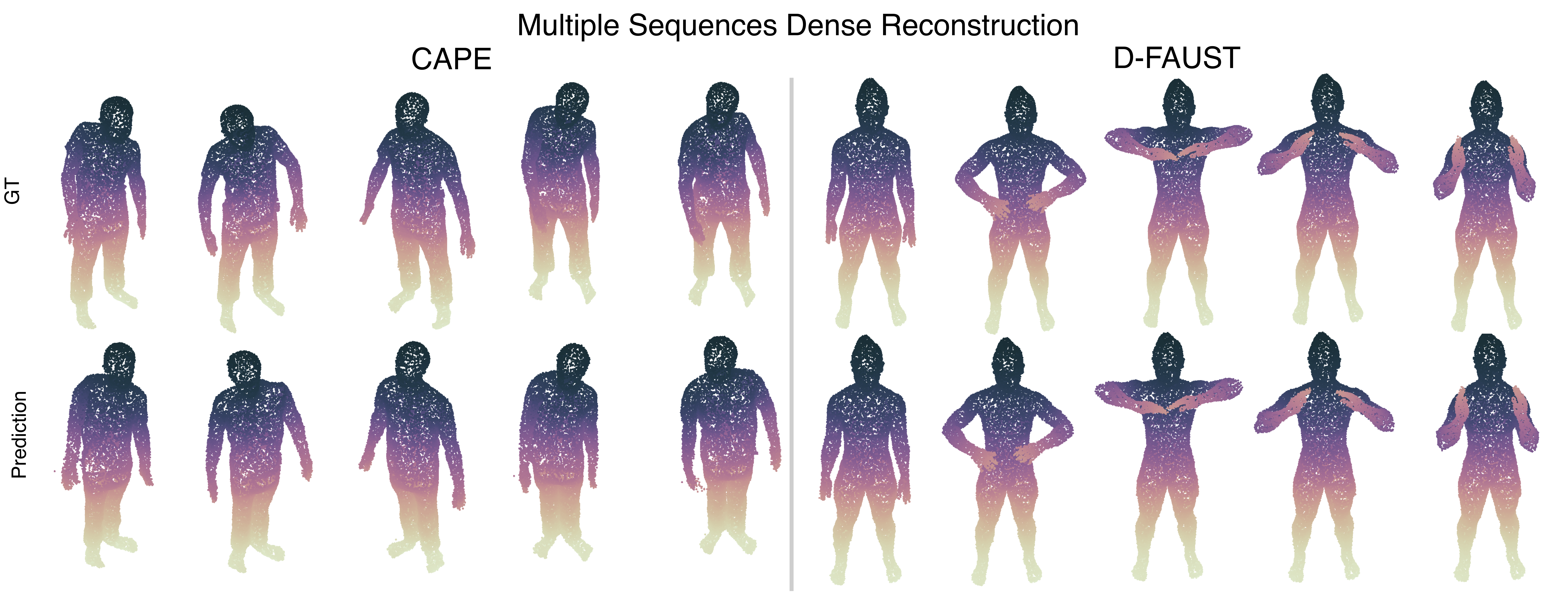}
    \caption{Results of time-consistent dense correspondence estimation for 3D point-cloud inputs from the CAPE~\cite{CAPE} (top) and DFAUST~\cite{dfaust2017} (bottom) datasets. We show comparisons with the ground truth dense correspondence maps.}
    \label{fig:sparse_dense}
\end{figure*}

Given a dataset of multiple shape identities in various poses, but without any registration within or across identities, we jointly learn: (\emph{i}) the weights of a network that predicts globally consistent dense point correspondences, and (\emph{ii}) shape and pose latent embeddings. Instead of requiring per-sequence dense correspondences as supervision signal~\cite{Palafox21}, 

we use a self-supervision cycle, in similar spirit to ~\cite{Bhatnagar2020, Saito2021}, to learn the shape/pose latent spaces without any need for registration during training. 
Given an input posed shape, the forward network learns to deform it to its canonical t-pose. The backward network then learns to deform the t-posed shape back to the input posed shape. Together, forward and backward networks form an identity mapping, and the distance between predicted and input shapes is used as supervision signal. Our only requirement on the training dataset is that each shape identity should be observed in a canonical pose (e.g., T-pose), which is satisfied on a variety of datasets~\cite{Deform4D,CAPE,MANO17}). 

At test time, given new observations of an unseen identity, the weights of the network are frozen and shape and pose embedding vectors are jointly optimised by minimizing the same cycle-consistency loss.

\subsection{Shape/Pose Network and Embeddings}
To circumvent the lack of correspondences we use a self-supervision cycle that conducts a bi-directional mapping between input and canonical poses. Our model is composed of two networks, both conditioned on the shape and pose latents (see Section~\ref{fig:self_cycle}). Given an input posed shape, the forward network predicts a dense deformation field $\delta \tilde{x}$ that maps it onto the canonical t-pose while the backward network learns to deform the canonical shape back to the input pose. 

Figure~\ref{fig:self_cycle} shows an overview of the self-supervision cycle. Forward and backward networks are implemented as auto-decoders, with parameters $\theta_a$ and $\theta_b$, and EdgeConv layers that predict dense deformation fields.

\noindent \textbf{Forward Network:} 
The forward network learns a geometric-aware deformation field  that deforms the input posed shape to its canonical t-pose shape. More formally, the input to the network is a set of $N$ observed points on the surface of the shape associated with the $f^{th}$ frame and $c^{th}$ identity, $X^{cf} = \{x_i^{cf} \}^{N}_{i=1} \in \mathbb{R}^3$ to which we apply positional encoding $\gamma(\cdot)$~\cite{NeRF}. The forward network forms a dynamic graph $\mathcal{G}^l$, via k-NN, conditional on a learnable per-frame $D_p$-dimensional latent pose code $p_f$, and a fixed $D_s$-dimensional latent shape code $s_c$.
A series of EdgeConv layers are applied, each using the output features of the previous layer to re-estimate the graph $\mathcal{G}^l$ (see~\ref{back:edge_conv}). Per-point pooling is performed after the last layer and a shallow MLP is applied to predict dense point-wise deformations $\{\delta x^{cf}_i \}^{N}_{i=1} \in \mathbb{R}^3$.
\begin{align} 
    f^{a}_{\theta_a}: \mathbb{R}^{N \times 51} \times \nonumber \mathbb{R}^{D_s} \times \mathbb{R}^{D_p} &\rightarrow \mathbb{R}^{N \times 3} \\ 
f^{a}_{\theta_a}(\gamma(x_i), s_c, p_f) &= \delta \tilde{x}.
    \label{eq:forward}
\end{align} 
\noindent\textbf{Backward Network:}
The backward network learns the inverse dense deformation field that deforms the canonical t-pose shape to the posed shape: $\{\delta y^{cf}_i \}^{N}_{i=1} \in \mathbb{R}^3$. While the architecture is equivalent to the forward network, the parameters are not shared.
\begin{align} 
    f^{b}_{\theta_b}: \mathbb{R}^{N \times 51} \times \nonumber \mathbb{R}^{D_s} \times \mathbb{R}^{D_p} &\rightarrow \mathbb{R}^{N \times 3} \\ 
f^{b}_{\theta_b}(\gamma(\tilde{y_i}), s_c, p_f) &= \delta \tilde{y} 
    \label{eq:backward}
\end{align} 
\noindent \textbf{Losses:} Combining (\ref{eq:forward}) and (\ref{eq:backward}) we can close the cycle and have a self-supervision loop.
We define the loss $\mathcal{L}_{loop}$ as the $l_1$ distance between the shape predicted by the cycle and the input shape.
\begin{equation} \label{eq:cycle}
\begin{split}
    f^{a}_{\theta_a}(\gamma(x_i), s_c, p_f) = \delta \tilde{x_i} \\
     \tilde{y_i} = x_i + \delta \tilde{x_i}\\
    f^{b}_{\theta_b}(\gamma(\tilde{y_i}), s_c, p_f) = \delta \tilde{y_i} \\
     \tilde{x_i} = \tilde{y_i} + \delta \tilde{y_i}\\
\end{split}
\end{equation}
\begin{equation}
    \mathcal{L}_{loop}(\tilde{x}^{cf}_{i}, x^{cf}_{i}) = \parallel \tilde{x}^{cf}_{i} - x^{cf}_{i} \parallel 
\end{equation}
To prevent the network from learning an identity mapping by setting $\delta \tilde{x}$ and $\delta \tilde{y}$ to zero, we use an additional symmetric ICP loss that minimizes the distance between the ground truth t-pose shape points and the t-pose predicted by the forward network 
\begin{equation}
    \mathcal{L}_{icp}(\tilde{y}_i^{cf}) = \parallel \tilde{y}_i^{cf} - NN_{\mathcal{R}}(\tilde{y}_i^{cf}) \parallel^2_2
\end{equation}
Where $NN_{\mathcal{R}}(.)$ is a function that queries the nearest neighbour of a 3D point in the set $\mathcal{R}$ of input points.
Which we found important to prevent correspondence flipping in the presence of extreme motion (see ~\ref{subsec:ablation}).

To aid temporal smoothness we constrain the current and next pose latents to be close by adding an $l_1$ loss between them $\mathcal{L}_{lt}$.
Also we found it very useful to enforce temporal regularization for both networks output which we called spacial temporal loss $\mathcal{L}_{st}$.
\begin{equation}
    \mathcal{L}_{lt}(p_f, p_{f+1}) = \parallel p_f  - p_{f+1} \parallel \\
\end{equation}
\noindent Inspired by the intuition points in consecutive frames deform coherently ~\cite{ARAP07}.
And as we do not have access to dense correspondences, we use the input points and query the nearest neighbour of the next frame input points.
Then we use $l_1$ loss between the current network prediction and the prediction of the nearest points in the next frame.
\begin{equation}
\begin{split}
    \mathcal{L}^{a}_{st}(\tilde{x}_i^{cf}) = \parallel \delta \tilde{x}_i^{cf} -  \delta \tilde{x}_i^{c(f+1)} \parallel \\
    \delta \tilde{x}_i^{c(f+1)} = f^{a}_{\theta_a}(NN_{\mathcal{Q}^{a}}({x}_i^{c(f+1)})) \\
    \mathcal{L}^{b}_{st}(\tilde{y}_i^{cf}) = \parallel \delta \tilde{y}_i^{cf} -  \delta \tilde{y}_i^{c(f+1)} \parallel \\
    \delta \tilde{y}_i^{c(f+1)} = f^{b}_{\theta_a}(NN_{\mathcal{Q}^{b}}({y}_i^{c(f+1)})) \\
\end{split}
\label{eq:temp}
\end{equation}
Where $NN_{\mathcal{Q}}(.)$ is a function that queries the nearest neighbour of a 3D point in the set $\mathcal{Q}$ of input points.
Our final temporal regularization loss is:

\begin{equation}
    \mathcal{L}_{temp} = \mathcal{L}_{lt} + \mathcal{L}^{a}_{st} + \mathcal{L}^{b}_{st}
\end{equation}

Finally, we minimize this objective over all possible $F$ deformation fields across all shape identities $C$ with respect to the individual pose and shape codes $\{p_f\}^F_{f=1}$ and $\{s_c\}^C_{c=1}$ respectively and the forward and backward network weights $\theta_a, \theta_b$.
To enforce a compact pose and shape manifold we also regularize the both codes with $\sigma_p$ and \; $\sigma_s$.

\begin{multline}
    \argmin_{\theta_a, \theta_b, \{s_c\}_{c=1}^{C}, \{p_f\}_{f=1}^{F}} \overset{C}{\underset{\substack{c=1}}{\sum}} \; \overset{F}{\underset{\substack{f=1}}{\sum}} \; \overset{N}{\underset{\substack{i=1}}{\sum}}\; \mathcal{L}_{loop} + \lambda_{icp} \mathcal{L}_{icp} \\
    + \lambda_{temp} \mathcal{L}_{temp} + \frac{\| p_f \|^2_2}{\sigma^2_p} + \frac{\| s_c \|^2_2}{\sigma^2_s}
\end{multline}

\subsubsection{EdgeConv Layers}
\label{back:edge_conv}

In contrast to previous neural parametric models~\cite{Palafox21} that treat each data point independently, we exploit the power of edge convolutions in DGCNNs~\cite{YueWang18} to learn local geometric structure in deformed 3D point clouds.  EdgeConv~\cite{YueWang18} alleviates the lack of topology in point clouds by proposing a differential neural network module suitable for CNN-based high-level tasks. Local geometric structure is exploited by constructing a local neighbourhood graph and applying convolution-like operations on the edges in a dynamic setting, where the connectivity is learned and changes throughout the training. 
The learned feature space not only captures semantic relations within a local neighbourhood, but also between distant points, which is particularly useful in the context of deformations where different body parts might move synchronously.

Given a set of $N$ uniformly sampled 3D surface points, denoted by $X = \{x_1, . . . , x_N\} \subseteq \mathbb{R}^{\textsc{D}}$, where $\textsc{D}$ is the point cloud dimensionality, a directed graph $G = (V, E)$  of vertices $V$ and edges $E$, is first built to represent the local point cloud structure via k-nearest neighbours (k-NN).  In our setting, input points are represented by their 3D coordinates $x_i = (x_i ,y_i , z_i)$, but ${\textsc{D}}$ more generally refers to the dimensionality at each layer. Edge features are then defined as $e_{ij} = h_{\Theta}(x_i , x_j )$, where $h_{\Theta} : \mathbb{R}^{\textsc{D}} \times \mathbb{R}^{\textsc{D}} \xrightarrow[]{} \mathbb{R}^{{\textsc{D}}'}$ is a multi-layer perceptron (MLP) with learnable parameters $\Theta$. Channel-wise symmetric aggregation $\Box$ (e.g., $\sum$ or $max$) is then applied on the edge features associated with each vertex $i$ with its output given by
\begin{equation}
    x'_i = \underset{j:(i, j) \in \mathcal{E}}{\Box} h_{\Theta}(x_i, x_j). 
\end{equation}

Unlike GCNNs that work on a fixed input graph, EdgeConv  recomputes the graph neighbourhood structure at each layer. In practice, a pairwise distance matrix is computed in feature space, and the closest $k$ points are selected for each vertex point at each layer.

\begin{figure*}[t]
    \centering
    \includegraphics[width = 0.85 \linewidth]{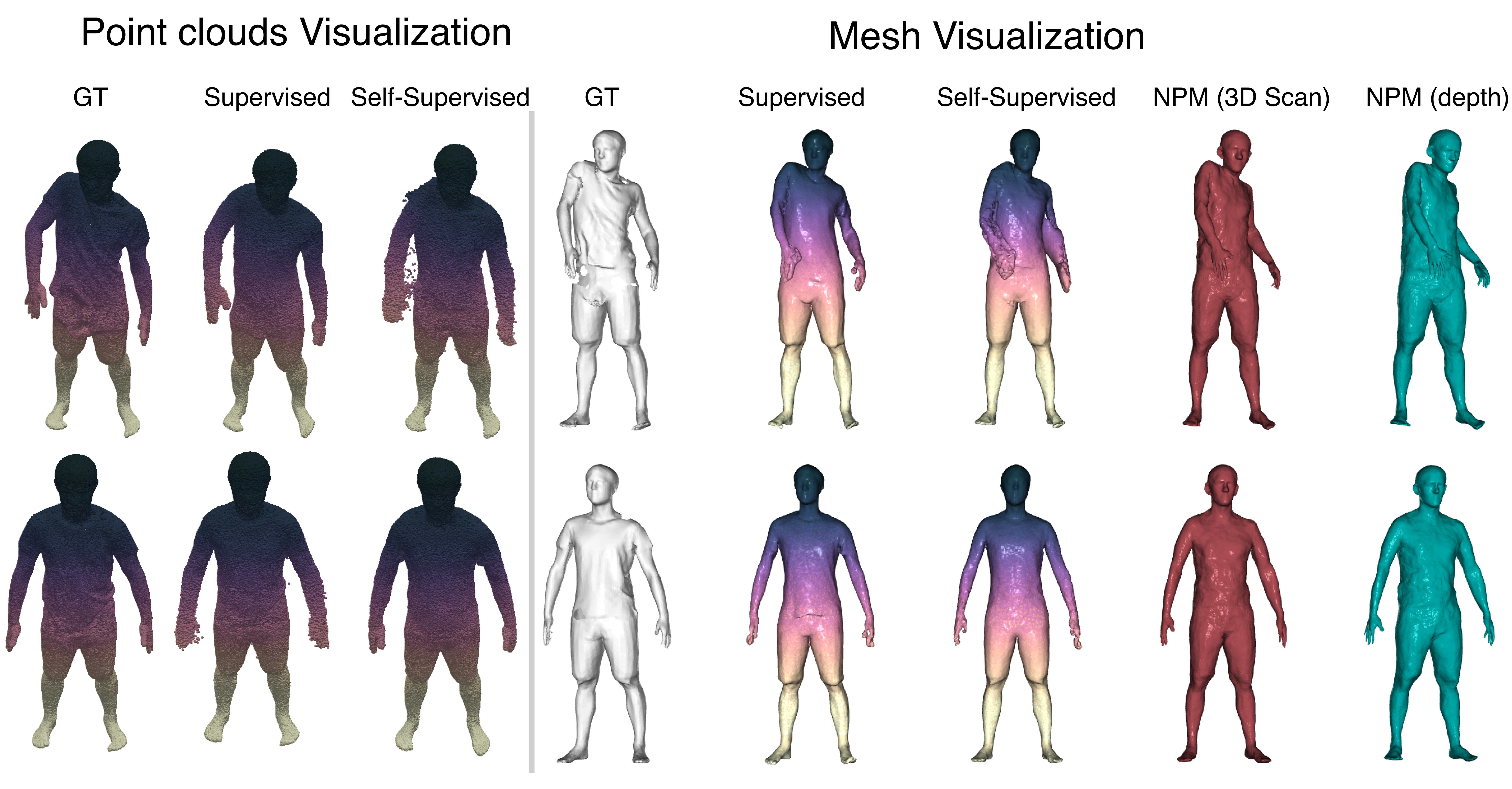}
    \caption{\textbf{Mesh Visualisation.}
    We sampled 300K points and use Poisson surface reconstruction ~\cite{kazhdan06poisson} to get meshes from our model.
    Left we have the reconstructed point clouds. Right we have the reconstructed meshes with Poisson.
    We use a supervised GNPM version of our model where we assume we have access to dense correspondences.
    And we also shows the result of evaluating NPM~\cite{Palafox21} on depth and complete 3D scans.
    As we can see that our method supervised/self-supervised can recover good quality meshes.
    }
    \label{fig:gnpm_poisson_mesh}
\end{figure*}

\subsection{Test Optimization}

Once the network weights and shape and pose latent embeddings have been learnt, they can be used to fit a new sequence to the model by optimizing for the per-identity shape and per-frame pose codes that best explain the observations. We use the pose network and minimize the cycle consistency training objective with respect to the shape and pose latent codes $\{s_c\}^S_{c=1}, \{p_l\}^F_{f=1}$, keeping network weights fixed:

\begin{multline}
    \argmin_{\{s_c\}_{c=1}^{C}, \{p_f\}_{f=1}^{F}} \overset{C}{\underset{\substack{c=1}}{\sum}} \; \overset{F}{\underset{\substack{f=1}}{\sum}} \; \overset{N}{\underset{\substack{i=1}}{\sum}}\; \mathcal{L}_{loop} 
    + \frac{\| p_f \|^2_2}{\sigma^2_p} + \frac{\| s_c \|^2_2}{\sigma^2_s}
\end{multline}
The shape and pose codes are initialized to the mean of the respective embedding. When dealing with a sequence, the pose code for each frame can be initialized with the result for the previous frame. 

\subsection{Implementation Details}

\noindent \textbf{Forward and Backward Networks:}
For both forward and backward networks, we stack three EdgeConv layers with two MLPs to compute edge features. LeakyReLU with $0.2$ negative slope was used as the activation function and $max$ as the channel-wise aggregation function for all layers. The graph connectivity is updated for each layer, based on the learned features in previous ones. Features from each EdgeConv are concatenated and passed through the MLPs for the final prediction. The dimensionality of the EdgeConv features is set to ${\textsc{D}}=128$, except for the last layer where ${\textsc{D}}=256$. Shape and pose latent dimensionalities are set to $128$.
We use the Adam optimizer~\cite{AdamKingma2015} with learning rates of $1e-3$ for the shape/pose model and latent codes. In addition, we apply a learning rate scheduler with a decay factor of 0.5 every 30 epochs.
And use a cosine annealing our ICP loss weight $\lambda_{icp}$ with $1e^{-1}$ initial weight and $1e^{-2}$ minimum weight.
As for the regularization losses we use $5e^{-2}$ for temporal loss weight $\lambda_{temp}$.
We use shape and pose code regularization weights $\sigma_s=\sigma_p=1e-4$.
Despite the symbolic k-NN implementation allowing more point samples, we sample 1024 points per batch from each shape as input to the model. The latent codes are initialized randomly from a normal distribution.
We use the positional encoding proposed in~\cite{NeRF} to encode query points with 8 frequency bands.

\begin{figure}[t!]
    \centering
    \includegraphics[width = 1 \linewidth]{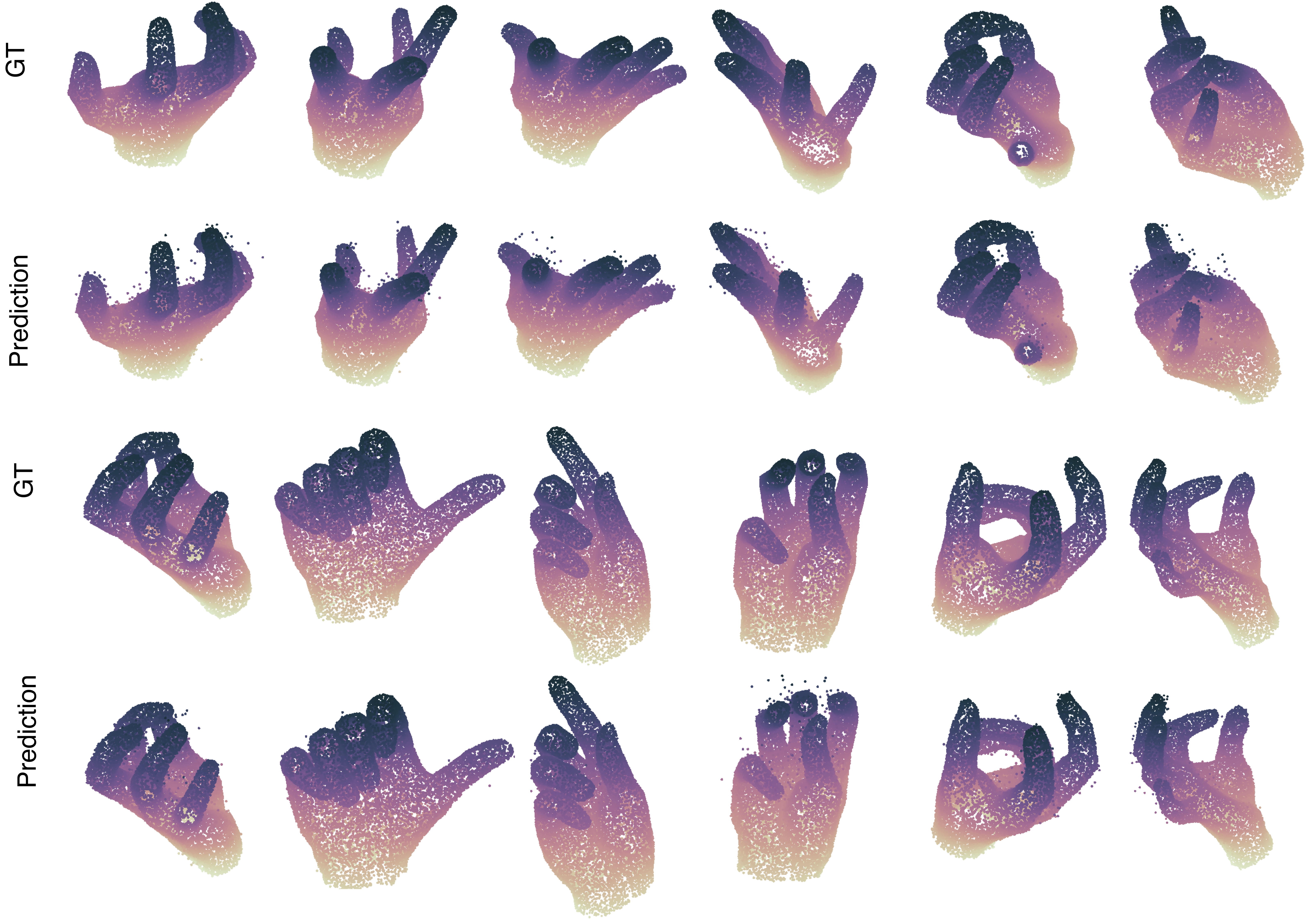}
    \caption{Results of time-consistent dense correspondence estimation for 3D point-cloud inputs from the MANO~\cite{MANO17} dataset. We show ground truth dense correspondences for comparison.}
    \label{fig:mano}
\end{figure}

\noindent \textbf{Efficient k-NN Implementation:}
We use the KeyOps~\cite{KeyOps21, feydy2020fast} python package to symbolically implement the k-NN algorithm. In the supplementary material we provide a comparison with a naive dense k-NN implementation and show a tenfold improvement in terms of training time.

\begin{figure}[h!]
    \centering
    \includegraphics[width = 1.0 \linewidth]{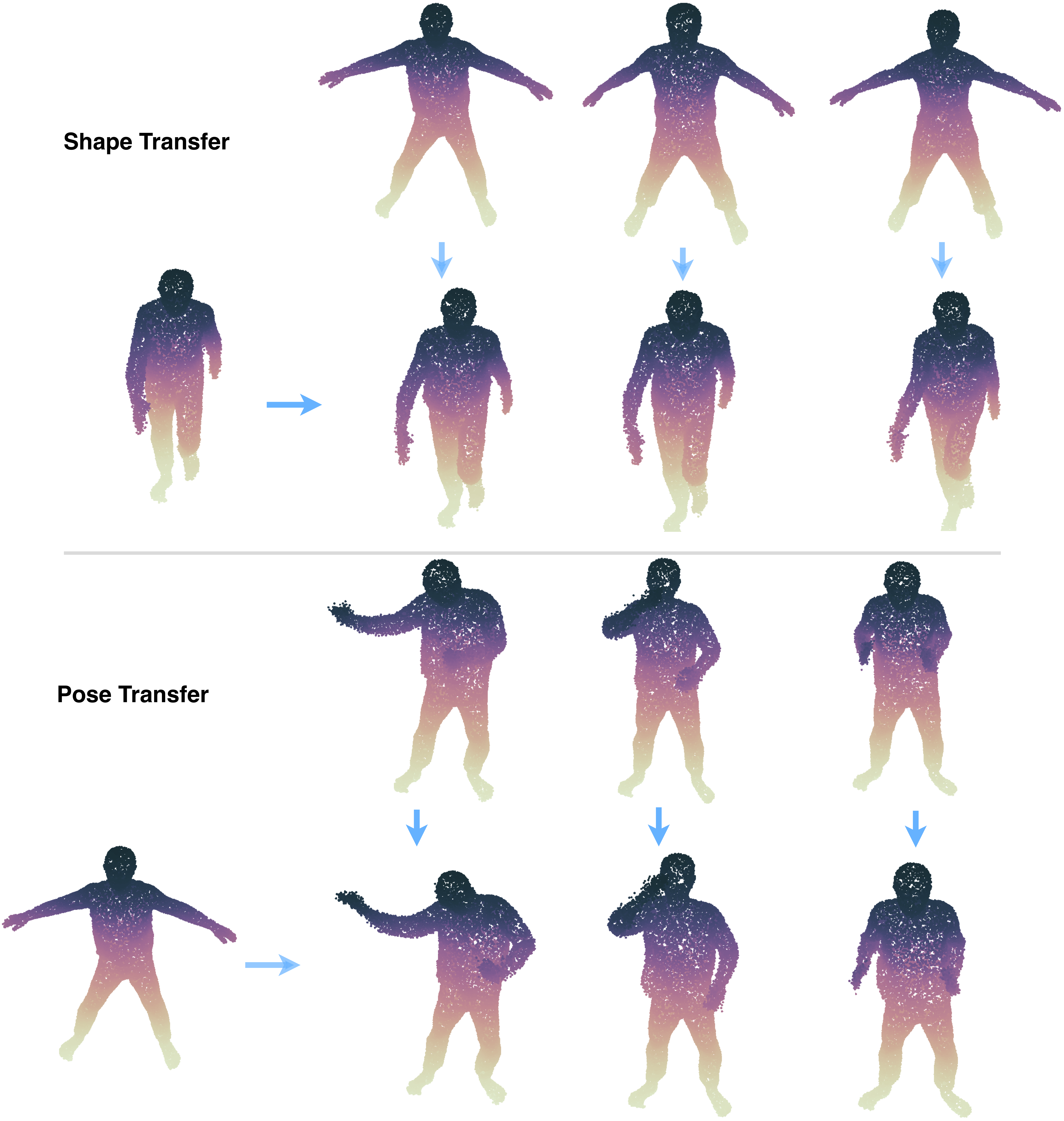}
    \caption{\textbf{Shape and pose transfer in latent space.}
    We can transfer a given identity to a posed shape (shape transfer). In addition, given a source identity in different poses, we can repose a target identity with the poses of the source (pose transfer).
    }
    \label{fig:sp_transfer}
\end{figure}

\begin{figure}[h!]
    \centering
    \includegraphics[width = 1 \linewidth]{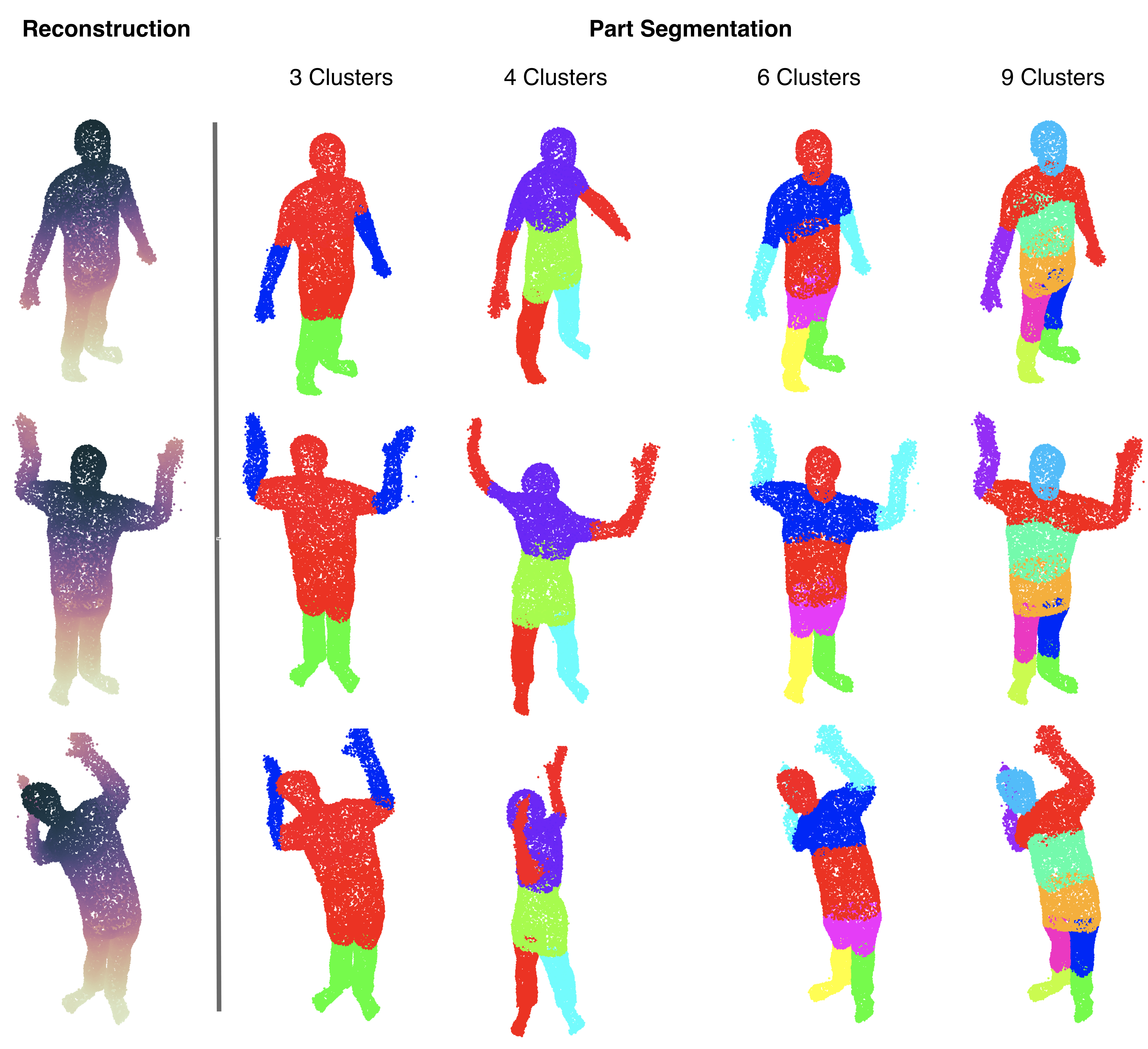}
    \caption{\textbf{Unsupervised 3D part segmentation with clustering on learned feature space with EdgeConv.}
    In left column the clusters are computed with Kmeans using 7 clusters on the features of first EdgConv layer.
    And in right column we experiment with different number of clusters and observed increasing the number of cluster leads to less semantically meaningful parts.
    }
    \label{fig:deformed_clust}
\end{figure}

\begin{figure}[!h]
    \centering
    \includegraphics[width = 0.9\linewidth]{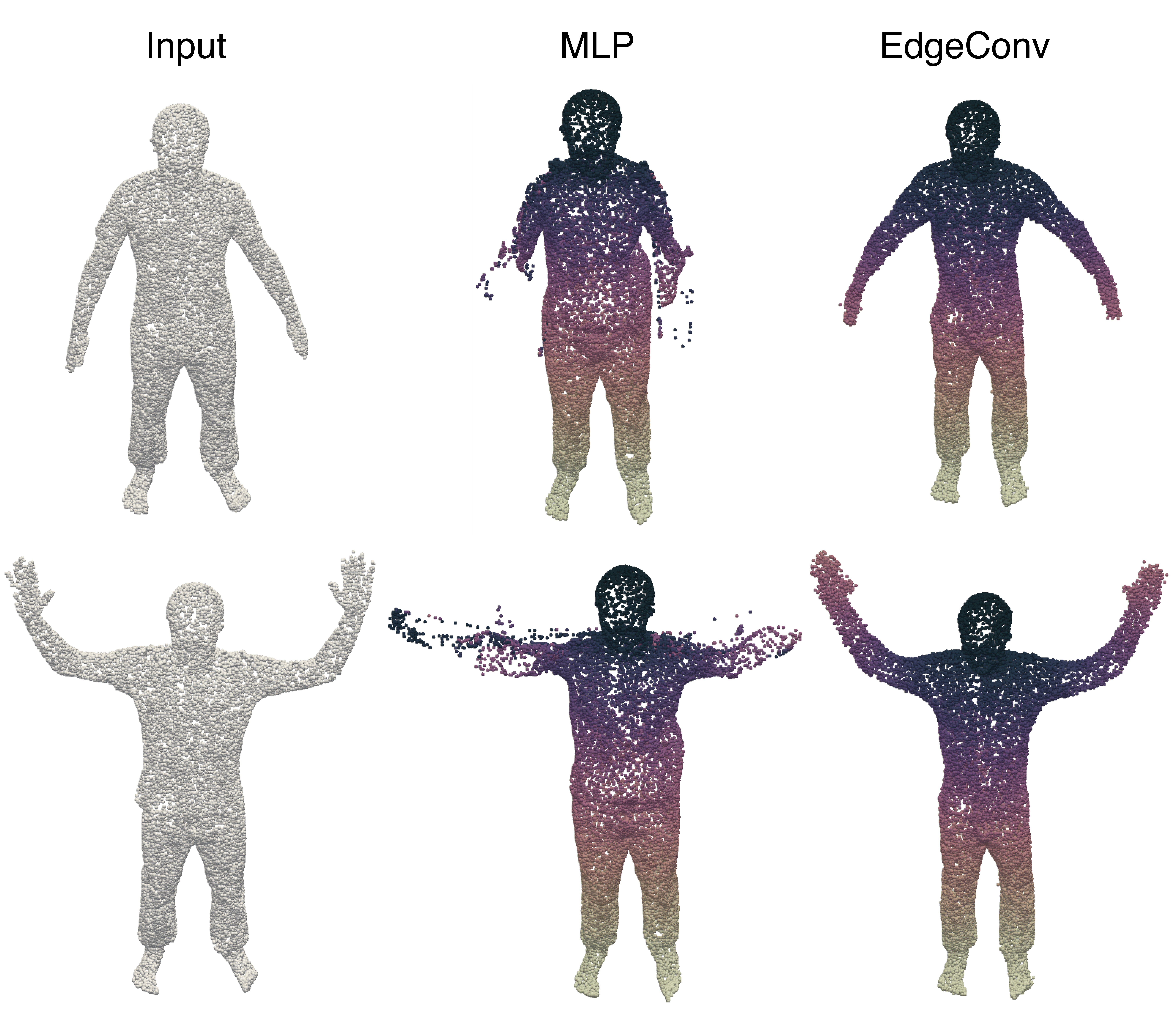}
    \caption{Qualitative evaluation of geometric inductive bias on CAPE~\cite{CAPE} datasets.
    We replace all EdgeConv layers with MLPs following the NPM~\cite{Palafox21} architecture choices for the MLPs.
    As we can see the MLP-based architecture struggles to model the deformation of the arms while the EdgConv based architecture does not suffer from this problem.}
    \label{fig:edge_mlp}
\end{figure}

\section{Experimental Evaluation}
\noindent \textbf{Datasets:}
We evaluate our model on real-world scans from the CAPE~\cite{CAPE} and D-FAUST~\cite{dfaust2017} datasets, which provide real clothed humans and their corresponding SMPL+D registration. We train on 31 different posed shapes from 35 different distinct identities~\cite{CAPE} and test on a total of 4 test sequences from 4 unseen characters. We also learn a hand model from the MANO ~\cite{MANO17} dataset.\\

\noindent\textbf{Evaluation metrics:}
We measure reconstruction and 4D tracking performance following the established per-frame evaluation protocol~\cite{Mescheder18, Palafox21}.
The $l_2$ Chamfer distance (C-$l_2$) offers a measure combining the accuracy and completeness of the reconstructed surface.
We also use End-Point Error (EPE), which measures keyframe-to-frame deformation $l_2$ distance between predicted and ground truth as in ~\cite{bozic2021neuraldg}. We visualise corresponding points with the same colour. 

\begin{table*}[!h]
    \begin{center}
    \resizebox{0.7\textwidth}{!}{\begin{tabular}{lccccll}
    \cline{1-5}
    \textbf{Method} & Input &\textbf{EPE} $\downarrow$ & \textbf{C-$l_2$} $(\times 10^{-3})$ $\downarrow$ &  \\ \cline{1-5}
    NPM~\cite{Palafox21}  & \emph{3D scan}  &\textbf{0.231}&    {0.019}   &  \\
    GNPM (K 10)  & \emph{3D scan}  & {0.287}&    \textbf{0.0122}   &  \\ \cline{1-5}
    \end{tabular}}
    \end{center}
    \caption{Comparison with NPM~\cite{Palafox21} on real scans of CAPE~\cite{CAPE}. 
    Since GNPM (our approach) requires full scans as input, we also evaluated NPM~\cite{Palafox21} on complete input scans for a fair comparison. Note that, in contrast to our method,  NPM~\cite{Palafox21} can take partial observations (depth maps) as input at test time. However, it requires known dense correspondences at training time, unlike our approach.}
    \label{tab:gnpm}
\end{table*}

\subsection{Evaluation on CAPE Dataset}

\noindent \textbf{Reconstruction and unsupervised dense correspondences:}
We compare our model to the state of the art on the CAPE ~\cite{CAPE} dataset which provides ground truth dense correspondences.
We compare with NPM~\cite{Palafox21} (the most related approach to ours), although it requires dense correspondences at training time, unlike our approach.

Figure~\ref{fig:sparse_dense} shows how our method is capable of reconstructing the deformed shapes and estimating temporally consistent point correspondences accurately. Despite only using 1024 points at training time, we can densify the points during test time by sampling new points and combining the result to get a dense deformed version of each shape.
We use this property to further evaluate the quality of the deformation and use Poisson surface reconstruction~\cite{kazhdan06poisson} on 300K points to get surface reconstruction.

Figure~\ref{fig:gnpm_poisson_mesh} shows a visualisation of the meshes obtained after reconstructing with our method followed by Poisson reconstruction~\cite{kazhdan06poisson}. We note that Poisson surface reconstruction ~\cite{kazhdan06poisson} requires surface normals which we estimate. For fair comparison we sample 300K GT points and run them through Poisson reconstruction to obtain the GT meshes. We also visualise the results obtained by NPM~\cite{Palafox21} with two types of inputs: depth images and 3D scans.
As we can see our method can obtain reasonable meshes.

In addition, we trained a supervised version GNPM where GT dense correspondences were given at training time (similarly to NPM). For this model we only use the backward network and we learn to disentangle shape and pose latents with the same network. Figure~\ref{fig:gnpm_poisson_mesh} shows visualisations of   the predicted dense point clouds and reconstructed meshes with this 'supervised' version of our model.

Table~\ref{tab:gnpm} shows a numerical evaluation against NPM~\cite{Palafox21} on the CAPE test set.
It is important to note that our approach GNPM can only be used with complete scans as input. For that reason we also evaluated NPM~\cite{Palafox21} using complete input scans even though this method can take partial observations (depth images) as input. But since ours cannot, we evaluated on complete input scans for fairness.

Table~\ref{tab:gnpm} shows our method achieves better Chamfer distance (C-$l_2$) while having a comparable End-Point Error (EPE) to NPM~\cite{Palafox21}, which is trained with ground truth dense correspondences, while our method works without them.

GNPM can find the shape and pose latents more efficiently during test time. While NPM~\cite{Palafox21} report that optimizing over an input sequence of 100 frames takes approximately 4 hours on a GeForce RTX 3090, our approach takes 20 minutes on a GeForce RTX 3080 on a similar length sequence.

\begin{table}[!h]
    \begin{center}
    \begin{tabular}{lcccll}
    \cline{1-4}
    \textbf{Method} & Input &\multicolumn{1}{l}{\textbf{EPE} $\downarrow$} &     \multicolumn{1}{l}{\textbf{C-$l_2$} $(\times 10^{-3})$ $\downarrow$} &  &  \\ \cline{1-4}
    ONet 4D~\cite{Mescheder18}      &  \emph{3D scan}  &     -     &         0.028        &  &  \\
    OFlow~\cite{OccFlowNiemeyer19ICCV}      &  \emph{3D scan}  &     -     &         0.031        &  &  \\
    GNPM (K 10)  & \emph{3D scan}  &\textbf{0.253}&    \textbf{0.0094}   &  &  \\ \cline{1-4}
    \end{tabular}
    \end{center}
    \caption{Comparison with state-of-the-art methods on real scans of the DFAUST~\cite{dfaust2017} dataset. Since OFlow~\cite{OccFlowNiemeyer19ICCV} works only on sequences of up to 17 frames, we report the average over sub-sequences of such length.}
    \label{tab:dfaus_comparison}
\end{table}

\subsection{Evaluation on DFAUST Dataset}

Table~\ref{tab:dfaus_comparison} shows a quantitative evaluation on the DFAUST dataset~\cite{dfaust2017}. All methods take 3D scans as input and our approach results in the lowest Chamfer distance error. We take the results for ONet~\cite{Mescheder18} and OFlow~\cite{OccFlowNiemeyer19ICCV_2} directly from the publications.

\subsection{Qualitative evaluation on Mano Dataset}

\textbf{Hand reconstruction.}
We demonstrate GNPMs ability to work on the MANO hand dataset~\cite{MANO17}.
Figure~\ref{fig:mano} shows our GNPM results showing its ability to model complex deformations of the hand and to establish temporally consistent and dense correspondences. Our comparison with ground truth correspondences shows that GNPM can infer high quality dense 4D temporal tracking, despite the challenging deformations shown by hands. 

\subsection{Latent Applications}
\noindent\textbf{Shape and Pose Transfer:}
Disentangling shape and pose spaces allows us to transfer shape and/or pose between identities. Given an input identity in a specific pose, we can map the same pose to new identities. Alternatively, we can fix the identity and transfer pose latents. Examples of both are shown in Fig.~\ref{fig:sp_transfer}.

\subsection{Unsupervised 3D Part Segmentation}

In this section we explore the potential of the features learnt by the EdgeConv network. Figure~\ref{fig:deformed_clust}, shows the result of performing unsupervised clustering on the learnt features for each frame independently, which leads to consistent part segmentations across identities and poses.
We can observe that by grouping  points according to their features we discover \textit{locally consistent} groups, to give a semantically meaningful part segmentation. We experimented with different cluster sizes as shown in Fig.~\ref{fig:deformed_clust}. 

\subsection{Ablation Studies}
\label{subsec:ablation}
\noindent\textbf{Geometric Inductive Bias:}
To evaluate the our architecture choices, we replace all EdgeConv layers with MLP-based forward and backward networks using NPM~\cite{Palafox21} pose model architecture choices and trained the networks on CAPE~\cite{CAPE} dataset.
The MLP-based self-supervised architecture infers the deformation field independently for each point without considering local neighbourhood structure.
As shown in Table ~\ref{tab:edge_mlp}, we found using EdgeConv layers is advantageous to learning deformations.
Figure ~\ref{fig:edge_mlp} shows qualitative evaluation.
MLP based architecture struggles with modelling arms and can not learn a deformation field that fully reconstructs the shape.
On the other hand, the EdgConv based architecture does not suffer from this problem. We conclude this inductive bias is important in learning deformation fields without dense correspondences.
And removing this bias results in a degenerate performance.\\

\begin{table}[!h]
    \begin{center}
    \resizebox{0.52\textwidth}{!}{\begin{tabular}{lccll}
    \cline{1-3}
    \textbf{Method} & \multicolumn{1}{l}{\textbf{EPE} $\downarrow$} &     \multicolumn{1}{l}{\textbf{C-$l_2$} $(\times 10^{-3})$ $\downarrow$} &  &  \\ \cline{1-3}
     Self-supervision (MLP)     & 0.515 &  0.0635  &  &  \\
     Self-supervision (EdgConv)     &\textbf{0.287} &  \textbf{0.0122}   &  &  \\ \cline{1-3}
    \end{tabular}}
    \end{center}
    \caption{Evaluation of geometric inductive bias.
    We use NPM~\cite{Palafox21} MLP architecture choices and replace all EdgeConv layer with MLPs.
    As we can see without considering the local neighbourhood around the points the model fail to learn deformation field.}
    \label{tab:edge_mlp}
\end{table}

\noindent\textbf{ICP Loss:}
We found that for frames with extreme deformation, the forward network has some difficulties dealing with points from similar parts like hands that change sides with respect to the canonical t-pose see supplementary materials.
Using an asymmetric ICP loss leads to flipping in the prediction of the forward network.
The backward network can handle this to some extent but as we are enforcing a geometric constraint, it fails to completely reverse points flipping resulting in a knot in the waist of the shape. Using a symmetric ICP loss where the nearest neighbours distance is computed from both source and target is important in learning extreme deformation without dense correspondences.\\

\section{Limitations}
While GNPM demonstrates potential for learning disentangled pose and shape space without dense correspondences, it still struggles with modelling fine details.
For instance, in modelling human deformation often the hand is not modelled in great detail.
A potential reason for that could be our uniform sampling approach. Denser sampling in areas like hands and faces could lead to better estimates of detailed deformations. Finally, our method is restricted to full meshes at test time. Future work includes extending our method to deal with partial observations such as depth maps.

\section{Conclusion}

We have proposed Geometric Neural Parametric Models (GNPM), a learned parametric model that takes as input point-clouds of different identities performing complex motions and disentangles shape/identity and deformations by mapping each pose to its canonical t-pose configuration and learning dense time-consistent correspondences. In addition, our model learns disentangled shape and pose embeddings which can later be used for interpolation, editing or transfer. 

In contrast to previous learnt neural parametric models such as NPMs~\cite{Palafox21}, our model does not require ground truth known correspondences at training time, which are costly to obtain. We exploit the concept of cycle-consistency to establish correspondences and show results on real-world scans from the CAPE~\cite{CAPE}, DFAUST~\cite{dfaust2017} and MANO~\cite{MANO17} datasets. Our network uses edge convolutions to extract features, which can be further exploited for downstream tasks such as unsupervised part segmentation. While our current approach cannot deal with partial scans from depth input images we envisage a completion network as future work. 


\section*{Acknowledgements}

Research presented here has been supported by the UCL Centre for Doctoral Training in Foundational AI under UKRI grant number EP/S021566/1.

{\small
\bibliographystyle{ieee_fullname}
\bibliography{egbib}

\begin{thebibliography}{10}\itemsep=-1pt

\bibitem{Scape05}
Dragomir Anguelov, Praveen Srinivasan, Daphne Koller, Sebastian Thrun, Jim
  Rodgers, and James Davis.
\newblock Scape: shape completion and animation of people.
\newblock {\em ACM Trans. Graph}, 24:408--416, 2005.

\bibitem{Bhatnagar2020}
Bharat~Lal Bhatnagar, Cristian Sminchisescu, Christian Theobalt, and Gerard
  Pons-Moll.
\newblock {LoopReg: Self-supervised learning of implicit surface
  correspondences, pose and shape for 3D human mesh registration}.
\newblock {\em Advances in Neural Information Processing Systems}, (NeurIPS),
  2020.

\bibitem{dfaust2017}
Federica Bogo, Javier Romero, Gerard Pons-Moll, and Michael~J. Black.
\newblock Dynamic {FAUST}: {R}egistering human bodies in motion.
\newblock In {\em IEEE Conf. on Computer Vision and Pattern Recognition
  (CVPR)}, July 2017.

\bibitem{bozic2021neuraldg}
Alja{\v{z}} Bo{\v{z}}i{\v{c}}, Pablo Palafox, Michael Zollh{\"o}fer, Justus
  Thies, Angela Dai, and Matthias Nie{\ss}ner.
\newblock Neural deformation graphs for globally-consistent non-rigid
  reconstruction.
\newblock {\em CVPR}, 2021.

\bibitem{KeyOps21}
Benjamin Charlier, Jean Feydy, Joan~Alexis Glaunès, François-David Collin,
  and Ghislain Durif.
\newblock Kernel operations on the gpu, with autodiff, without memory
  overflows.
\newblock {\em Journal of Machine Learning Research}, 22(74):1--6, 2021.

\bibitem{Eisenberger21}
Marvin Eisenberger, David Novotn{\'{y}}, Gael Kerchenbaum, Patrick Labatut,
  Natalia Neverova, Daniel Cremers, and Andrea Vedaldi.
\newblock Neuromorph: Unsupervised shape interpolation and correspondence in
  one go.
\newblock In {\em {IEEE} Conference on Computer Vision and Pattern Recognition,
  {CVPR} 2021, virtual, June 19-25, 2021}, pages 7473--7483. Computer Vision
  Foundation / {IEEE}, 2021.

\bibitem{feydy2020fast}
Jean Feydy, Joan Glaun{\`e}s, Benjamin Charlier, and Michael Bronstein.
\newblock Fast geometric learning with symbolic matrices.
\newblock {\em Advances in Neural Information Processing Systems}, 33, 2020.

\bibitem{Joo15}
Hanbyul Joo, Tomas Simon, and Yaser Sheikh.
\newblock Total capture: {A} 3d deformation model for tracking faces, hands,
  and bodies.
\newblock In {\em 2018 {IEEE} Conference on Computer Vision and Pattern
  Recognition, {CVPR} 2018, Salt Lake City, UT, USA, June 18-22, 2018}, pages
  8320--8329. Computer Vision Foundation / {IEEE} Computer Society, 2018.

\bibitem{kazhdan06poisson}
Michael Kazhdan, Matthew Bolitho, and Hugues Hoppe.
\newblock Poisson surface reconstruction.

\bibitem{AdamKingma2015}
Diederik~P. Kingma and Jimmy~Lei Ba.
\newblock {Adam: A method for stochastic optimization}.
\newblock {\em 3rd International Conference on Learning Representations, ICLR
  2015 - Conference Track Proceedings}, pages 1--15, 2015.

\bibitem{FLAME17}
Tianye Li, Timo Bolkart, Michael~J. Black, Hao Li, and Javier Romero.
\newblock Learning a model of facial shape and expression from {4D} scans.
\newblock {\em ACM Transactions on Graphics}, 36(6):194:1--194:17, Nov. 2017.
\newblock Two first authors contributed equally.

\bibitem{Deform4D}
Yang Li, Hikari Takehara, Takafumi Taketomi, Bo Zheng, and Matthias Nießner.
\newblock 4dcomplete: Non-rigid motion estimation beyond the observable
  surface.
\newblock {\em arXiv preprint arXiv:2105.01905}, 2021.

\bibitem{SMPL15}
Matthew Loper, Naureen Mahmood, Javier Romero, Gerard Pons-Moll, and Michael~J.
  Black.
\newblock {SMPL}: A skinned multi-person linear model.
\newblock {\em ACM Trans. Graphics (Proc. SIGGRAPH Asia)}, 34(6):248:1--248:16,
  Oct. 2015.

\bibitem{CAPE}
Qianli Ma, Jinlong Yang, Anurag Ranjan, Sergi Pujades, Gerard Pons{-}Moll, Siyu
  Tang, and Michael~J. Black.
\newblock Learning to dress 3d people in generative clothing.
\newblock In {\em 2020 {IEEE/CVF} Conference on Computer Vision and Pattern
  Recognition, {CVPR} 2020, Seattle, WA, USA, June 13-19, 2020}, pages
  6468--6477. Computer Vision Foundation / {IEEE}, 2020.

\bibitem{Mescheder18}
Lars~M. Mescheder, Michael Oechsle, Michael Niemeyer, Sebastian Nowozin, and
  Andreas Geiger.
\newblock Occupancy networks: Learning 3d reconstruction in function space.
\newblock In {\em {IEEE} Conference on Computer Vision and Pattern Recognition,
  {CVPR} 2019, Long Beach, CA, USA, June 16-20, 2019}, pages 4460--4470.
  Computer Vision Foundation / {IEEE}, 2019.

\bibitem{NeRF}
Ben Mildenhall, Pratul~P. Srinivasan, Matthew Tancik, Jonathan~T. Barron, Ravi
  Ramamoorthi, and Ren Ng.
\newblock Nerf: Representing scenes as neural radiance fields for view
  synthesis.
\newblock 2020.

\bibitem{Newcombe15}
Richard~A. Newcombe, Dieter Fox, and Steven~M. Seitz.
\newblock Dynamicfusion: Reconstruction and tracking of non-rigid scenes in
  real-time.
\newblock In {\em {IEEE} Conference on Computer Vision and Pattern Recognition,
  {CVPR} 2015, Boston, MA, USA, June 7-12, 2015}, pages 343--352. {IEEE}
  Computer Society, 2015.

\bibitem{OccFlowNiemeyer19ICCV_2}
Michael Niemeyer, Lars Mescheder, Michael Oechsle, and Andreas Geiger.
\newblock {Occupancy flow: 4D reconstruction by learning particle dynamics}.
\newblock {\em Proceedings of the IEEE International Conference on Computer
  Vision}, pages 5378--5388, 2019.

\bibitem{OccFlowNiemeyer19ICCV}
Michael Niemeyer, Lars~M. Mescheder, Michael Oechsle, and Andreas Geiger.
\newblock Occupancy flow: 4d reconstruction by learning particle dynamics.
\newblock In {\em 2019 {IEEE/CVF} International Conference on Computer Vision,
  {ICCV} 2019, Seoul, Korea (South), October 27 - November 2, 2019}, pages
  5378--5388. {IEEE}, 2019.

\bibitem{Palafox21}
Pablo~R. Palafox, Aljaz Bozic, Justus Thies, Matthias Nie{\ss}ner, and Angela
  Dai.
\newblock Npms: Neural parametric models for 3d deformable shapes.
\newblock {\em ICCV 2021}.

\bibitem{DeepSDF}
Jeong~Joon Park, Peter Florence, Julian Straub, Richard~A. Newcombe, and Steven
  Lovegrove.
\newblock Deepsdf: Learning continuous signed distance functions for shape
  representation.
\newblock In {\em {IEEE} Conference on Computer Vision and Pattern Recognition,
  {CVPR} 2019, Long Beach, CA, USA, June 16-20, 2019}, pages 165--174. Computer
  Vision Foundation / {IEEE}, 2019.

\bibitem{Paschalidou2021CVPR}
Despoina Paschalidou, Angelos Katharopoulos, Andreas Geiger, and Sanja Fidler.
\newblock Neural parts: Learning expressive 3d shape abstractions with
  invertible neural networks.
\newblock In {\em Proceedings IEEE Conf. on Computer Vision and Pattern
  Recognition (CVPR)}, June 2021.

\bibitem{PaysanKARV09}
Pascal Paysan, Reinhard Knothe, Brian Amberg, Sami Romdhani, and Thomas Vetter.
\newblock A 3d face model for pose and illumination invariant face recognition.
\newblock In Stefano Tubaro and Jean{-}Luc Dugelay, editors, {\em Sixth {IEEE}
  International Conference on Advanced Video and Signal Based Surveillance,
  {AVSS} 2009, 2-4 September 2009, Genova, Italy}, pages 296--301.

\bibitem{Wang19}
Stylianos Ploumpis, Haoyang Wang, Nick~E. Pears, William A.~P. Smith, and
  Stefanos Zafeiriou.
\newblock Combining 3d morphable models: {A} large scale face-and-head model.
\newblock pages 10934--10943, 2019.

\bibitem{MANO17}
Javier Romero, Dimitrios Tzionas, and Michael~J. Black.
\newblock Embodied hands: Modeling and capturing hands and bodies together.
\newblock {\em ACM Transactions on Graphics, (Proc. SIGGRAPH Asia)}, 36(6),
  Nov. 2017.

\bibitem{Saito2021}
Shunsuke Saito, Jinlong Yang, Qianli Ma, and Michael~J. Black.
\newblock Scanimate: Weakly supervised learning of skinned clothed avatar
  networks.
\newblock pages 2886--2897, 2021.

\bibitem{ARAP07}
Olga Sorkine and Marc Alexa.
\newblock As-rigid-as-possible surface modeling.
\newblock In {\em Proceedings of EUROGRAPHICS/ACM SIGGRAPH Symposium on
  Geometry Processing}, pages 109--116, 2007.

\bibitem{YueWang18}
Yue Wang, Yongbin Sun, Ziwei Liu, Sanjay~E. Sarma, Michael~M. Bronstein, and
  Justin~M. Solomon.
\newblock Dynamic graph {CNN} for learning on point clouds.
\newblock {\em {ACM} Trans. Graph.}, 38(5):146:1--146:12, 2019.

\bibitem{Wu19}
Zonghan Wu, Shirui Pan, Fengwen Chen, Guodong Long, Chengqi Zhang, and
  Philip~S. Yu.
\newblock A comprehensive survey on graph neural networks.
\newblock {\em {IEEE} Trans. Neural Networks Learn. Syst.}, 32(1):4--24, 2021.

\bibitem{GHUMcvpr20}
Hongyi Xu, Eduard~Gabriel Bazavan, Andrei Zanfir, William~T. Freeman, Rahul
  Sukthankar, and Cristian Sminchisescu.
\newblock {GHUM} {\&} {GHUML:} generative 3d human shape and articulated pose
  models.
\newblock In {\em 2020 {IEEE/CVF} Conference on Computer Vision and Pattern
  Recognition, {CVPR} 2020, Seattle, WA, USA, June 13-19, 2020}, pages
  6183--6192, 2020.

\bibitem{ZuffiKJB16}
Silvia Zuffi, Angjoo Kanazawa, David~W Jacobs, and Michael~J Black.
\newblock 3d menagerie: Modeling the 3d shape and pose of animals.
\newblock In {\em Proceedings of the IEEE conference on computer vision and
  pattern recognition}, pages 6365--6373, 2017.

\end{thebibliography}
}

\end{document}